\documentclass{article}

     \usepackage[preprint]{neurips_2019}

\usepackage[utf8]{inputenc} % allow utf-8 input
\usepackage[T1]{fontenc}    % use 8-bit T1 fonts
\usepackage[colorlinks=false]{hyperref}      % hyperlinks
\usepackage{url}            % simple URL typesetting
\usepackage{booktabs}       % professional-quality tables
\usepackage{amsfonts}       % blackboard math symbols
\usepackage{nicefrac}       % compact symbols for 1/2, etc.
\usepackage{microtype}      % microtypography

\usepackage{caption}
\usepackage{graphicx}
\usepackage{amsmath}
\usepackage{booktabs}
\usepackage{algorithmicx}
\usepackage{algpseudocode}
\usepackage{amssymb}
\usepackage{subfigure}
\graphicspath{{Figures/}}
\usepackage{multirow}

\usepackage{wrapfig}
\usepackage[linesnumbered,vlined,ruled]{algorithm2e}
\usepackage{algpseudocode}
\SetKwInOut{Input}{input}
\SetKwInOut{Output}{output}

\newcommand{\tabincell}[2]{\begin{tabular}{@{}#1@{}}#2\end{tabular}}
\newcommand{\solver}{\texttt{LP-GNN}~}

\newcommand{\planModel}{\mathcal{M}}
\newcommand{\actions}{A}
\newcommand{\poss}{\alpha}
\newcommand{\Prop}{\mathbb{P}}
\newcommand{\domainM}{\mathcal{D}}

\newcommand{\planP}{\mathcal{P}}

\newcommand{\pre}{\mathsf{pre}}

\newcommand{\obs}{\sigma}
\newcommand{\learnP}{\mathcal{P}_L}

\newcommand{\va}{\mathbf{a}}
\newcommand{\ve}{\mathbf{e}}
\newcommand{\vp}{\mathbf{p}}
\newcommand{\vs}{\mathbf{s}}
\newcommand{\vP}{\mathbf{P}}
\newcommand{\vE}{\mathbf{E}}
\newcommand{\vS}{\mathbf{S}}
\newcommand{\vA}{\mathbf{A}}
\newcommand{\vactions}{\mathbf{\actions}}

\title{Representation Learning for Classical Planning from Partially Observed Traces}

\author{
Zhanhao Xiao$^1$, 
\bf Hai Wan$^1$, 
Hankui Hankz Zhuo$^1$, 
Jinxia Lin$^1$, 
Yanan Liu$^1$
\\
$^1$School of Data and Computer Science, Sun Yat-sen University, China
}

\begin{document}

\maketitle

\begin{abstract}
Specifying a complete domain model is time-consuming, which has been a bottleneck of AI planning technique application in many real-world scenarios. Most classical domain-model learning approaches output a domain model in the form of the declarative planning language, such as STRIPS or PDDL, and solve new planning instances by invoking an existing planner. However, planning in such a representation is sensitive to the accuracy of the learned domain model which probably cannot be used to solve real planning problems.
In this paper, to represent domain models in a vectorization representation way, we propose a novel framework based on graph neural network (GNN) integrating model-free learning and model-based planning, called \solver.
By embedding propositions and actions in a graph, the latent relationship between them is explored to form a domain-specific heuristics.
We evaluate our approach on five classical planning domains, comparing with the classical domain-model learner ARMS.
The experimental results show that the domain models learned by our approach are much more effective on solving real planning problems.
%with propositions and states represented as vertexes and actions as vectors. The state progresses by performing an action when the proposition-state graph progresses with the edge being updated by the action vector. For such a planning representation, we propose a novel framework to learn domain models from partially observed traces, called \solver. As the interpretation of propositions in a state is captured by edges, a new planning instance can be computed via the domain model learned.
%framework of learning domain models and planning based on graph neural network (GNN) from partially observed traces, called AML-GNN. To capture the relation between the propositions and the state, we build a GNN block where propositions and states are represented as vertexes and the interpretations are represented as edges. Every partially observed trace is captured in a recursive model consisting of GNN blocks and by updating the graph via an action vector, the state progresses. Due to the vectorization of propositions, the new states can be represented so that new planning instances can be solved in our framework.
%Furthermore, thanks to vectorizing states, we propose an approach to learning the heuristic function based on the neural network to assist planning.
%The results of the experiments on three well-known domains show that our approach outperforms the domain-model learner ARMS and the heuristic function learned is effective.
\end{abstract}

\section{Introduction}
AI planning techniques generally require domain experts to provide background knowledge about the dynamics of the planning domains. %, which have been applied in many real-world scenarios \citep{DBLP:books/daglib/0014222,rajan2013towards}
But Specifying a complete domain model is time-consuming, which has been a bottleneck of AI planning technique application in many real-world scenarios.
Taking an example of arranging production lines in a smart factory, there are a vast number of actions and predicates, it is difficult for humans to design an appropriate domain model that covers all actions.
%The increasing applications of AI planning techniques in the real world requires

However, most traditional domain-model learning approaches output a domain model in a kind of declarative planning language, such as STRIPS or PDDL, where the precondition and effects of actions are given in a declarative way.
With the learned domain models, a planner for the planning language is invoked to compute a plan to new planning problems.
But whether a plan can be found is sensitive to the accuracy of the learned domain model.
Once some critical effect of an action is not learned correctly, the error will accelerate with the plan growing, which finally leads that there is no plan to the goal computed.
One promising way is to keep away from learning domain models in the declarative language and to find a new representation to learn and then compute plans.
The new planning representation requires to satisfy at least the two following conditions: the state can be represented correctly; there is an effective way to compute a plan.
The former allows to represent a new planning instance and the latter is supposed to be efficient as possible, which requires a suitable heuristic function in the forward search planning.

Inspired by word embedding \citep{DBLP:journals/corr/abs-1301-3781} and knowledge graph embedding \citep{DBLP:conf/nips/BordesUGWY13} which have shown great success in natural language process and knowledge graphs,
it is constructive to represent propositions, states, and actions in the form of vectors.
To capture the relationship between propositions and states, we consider them jointly as vertexes with a real-number attribute vector in a graph where the interpretation of propositions in a state is captured by the directed edges.
%Furthermore, the relationship between states and actions may help us to find an appropriate heuristic function to guide planning.
%It suggests using graph neural network (GNN), which is a family of deep learning methods that operate on graphs.
%Due to its convincing performance and high interpretability, GNN has received much attention from researchers and is widely applied in graph analysis.
In this paper, we propose a novel learning and planning framework based on graph neural network (GNN), called \solver, taking the meaning of \textbf{L}earning to \textbf{P}lan based on GNN.
\solver integrates model-free learning from partially observed traces and model-based planning based on proposition-state graphs.
%the state progresses via the edges being updated by action vectors.
%
%The interpretation of propositions in a state is captured by the directed edges from the propositions to the state in the so-called proposition-state graph.
%By selecting an action to perform, the graph progresses with the edge being updated by the action vector.
%For such a planning representation, we propose a novel framework of learning domain models from partially observed traces, called \solver, taking the meaning of \textbf{L}earning to \textbf{P}lan based on GNN.
%Every partially observed plan trace is captured by a progression sequence of proposition-state graphs.
Due to the representation way of proposition-state graphs, a new state which has not occurred in all the plan traces can be denoted. It provides the possibility to generalize the planning system to handle new planning instances.

To improve the performance of planning, researchers in the planning community have proposed a number of heuristics, such as relaxed planning graph heuristics \citep{DBLP:journals/jair/HoffmannN01}, $h^m$ heuristics \citep{DBLP:journals/ai/BonetG01}, pattern database heuristics \citep{DBLP:conf/aips/Edelkamp02}, etc.
%These heuristics have their own advantages in different kinds of planning domains and there is not an approach to choose suitable heuristics for a new planning problem.
It suggests to choose an appropriate heuristic function for specific domains.
Based on proposition-state graphs, the relationship between states and actions are captured naturally, which may help us to find an appropriate heuristic function to guide planning.
%In this paper, thanks to vectorizing states and actions,
Therefore,
we propose an approach based on MLP to learn heuristic to guide selecting actions towards the goal state.
%The experiment results show that the heuristic learned is effective to find a plan.
To evaluate the learning and planning performances,
we compare \solver with the classical domain model learning system ARMS \citep{DBLP:journals/ai/YangWJ07} on five well-known planning domains and show that \solver outperforms ARMS and more robust on solving real planning problems.

%%\vspace{-2mm}
\section{Background}
%%\vspace{-2mm}
%A state $s$ in planning domain is a subset of proposition set $\Prop$.
Now we follow the notions of \citep{DBLP:conf/ijcai/FrancesRLG17} about classical planning.
We consider a set of propositions $\Prop$ and consider a state is a subset of $\Prop$ where the interpretation of proposition is given by the inclusion relation.
In other words, if $p_j \in s$, it means the proposition $p_j$ is true in the state $s$; otherwise, $p_j$ is false in $s$.
A classical planning problem is given as a tuple $\planModel = ( S,s_0,S_g,\actions,\poss,\gamma)$ where $S$ is a set of states, $s_0 \in S$ is an initial state, $S_g \subseteq S$ is a goal state set, $\actions$ is an action set, $\poss: S \longrightarrow 2^\actions$ is an applicable function, and $\gamma: S \times \actions \longrightarrow S$ is a state-transition function.
Intuitively, $\poss(s)$ indicates the actions applicable in state $s$ and $\gamma(s,a)$ represents the state resulting from performing the action $a$ in the state $s$.
A \emph{domain model} is a tuple $\domainM = (\poss,\gamma)$ and a \emph{planning instance} is a tuple $(s_0,S_g)$.
The solution to a classical planning problem is a \emph{plan} which is an action sequence $\pi = \langle a_1,...,a_n \rangle$ satisfying that there exists a state sequence $\langle s_0,...,s_n \rangle$ such that $\forall 0 \leq i \leq n, s_{i{+}1} = \gamma(s_i,a_{i{+}1})$, $a_{i{+}1} \in \poss(s_i)$ and $s_n\in S_g$.
%In this paper, we restrict the planning instance to be a pair of states $(s_0,s_g)$ where $s_0$ is an initial state and $s_g$ is a goal state.

A plan executed on a planning instance yields a plan trace, which is an alternating sequence of states and actions $\pi = \langle s_0,a_1,s_1,...,a_n,s_n \rangle$.
%To capture partial observation,
We suppose the initial state and goal state are fully observed and the intermediate states are not.
Formally, a partially observed plan traces is a sequence $\hat{\pi} = \langle s_0,a_1,\obs_1,...,a_n,s_n \rangle$, where $\obs_i \subseteq \Prop$.
Note that a proposition $p$ not in $\obs_i$ means to be either false or unobserved.
%A partially observed plan trace is a plan trace $\pi$ with an observation function $\obs: S_{\pi} \longrightarrow \Prop$ where $S_\pi$ is the set of states in $\pi$ and $s \subseteq \obs(s)$.
%When the state is partially observed, the propositions not in the state are either false or unobserved and
%the observation function $\obs$ indicates which propositions in every state are observed.
%Formally,
%for $p_j \in \Prop$, if $p_j \in s_i$, then we say $p_j$ is true in $s_i$, denoted by $p^i_j=1$; otherwise if $p_j \in \obs(s_i)\setminus s_i$, we say $p_j$ is false in $s_i$, denoted by $p^i_j=0$.
%We suppose the initial state and the goal state in the trace are totally observed.

We say a domain model $\domainM$ interprets a partially observed trace $\hat{\pi} = \langle s_0,a_1,\obs_1,...,a_n,s_n \rangle$ if $\langle a_1,...,a_n \rangle$ is a plan of the classical planning problem $\planModel = ( S,s_0,\{s_n\},\actions,\domainM)$ and
the yielded plan trace $\pi=\langle s_0,a_1,s_1,...,a_n,s_n \rangle$ satisfies that $\forall 0 \leq i \leq n$, $\obs_i \subseteq s_i$.
%each state in the state sequence induced by the plan includes the corresponding observation in $\hat{\pi}$.

A domain-model learning problem is a tuple $\learnP =(S,\actions,\Pi)$ where $\Pi$ is a set of partially observed plan traces.
A solution to the problem is a domain model which interprets all plan traces in $\Pi$.
%In this paper, we suppose actions always can be done and only focus on learning the state-transition function $\gamma$.

%%\vspace{-2mm}
\section{An Overview of Our Approach}
\vspace{-2mm}
In this section, we give an overview of our approach \textsc{LP-GNN}, which is based on GNN \cite{DBLP:journals/corr/abs-1806-01261}.
%The whole framework is shown in Figure \ref{fig:framework}.
The framework consists of two modules: the first one is the learning module which takes partially observed traces as input and outputs a domain model based on GNN and heuristics function; the second one is the planning module based on the learned heuristic function.

\begin{figure}[!htp]
\centering
\includegraphics[width=0.7\linewidth]{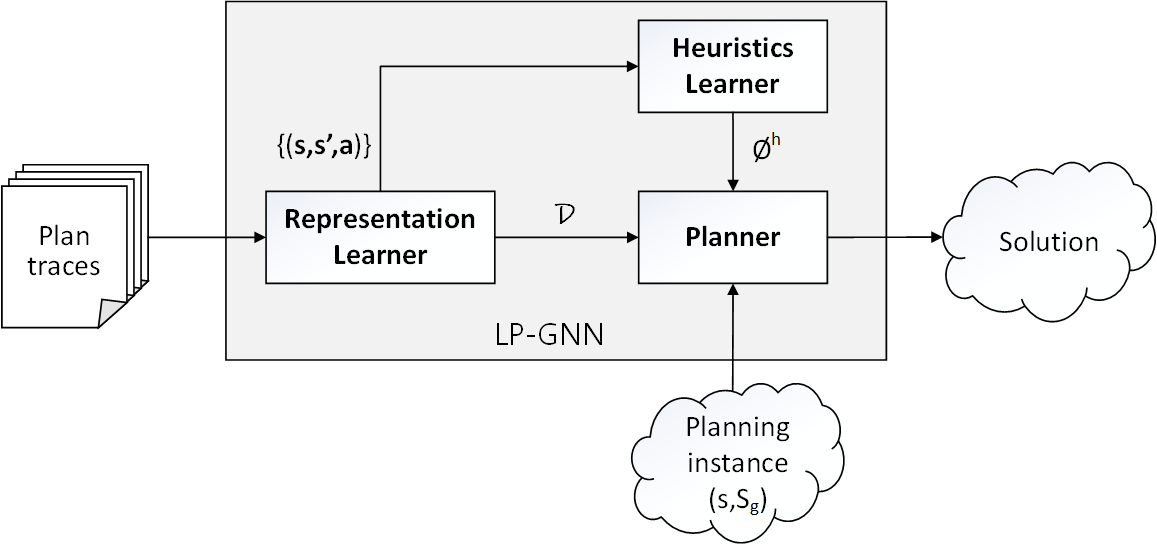}
\caption{An Overview of \solver}
\label{fig:framework}
\end{figure}

As states are not totally known in the partially observed traces, the hidden parts are required to be estimated in order to learn the domain model.
The estimation task is accomplished via a recurrent framework of graph network with a cost function evaluating the difference between estimated states and observed states.
When the framework converges, it outputs a set of sequences of completed states and every state is represented as a unique vector.
Also, it finally returns the vectorization representation of actions and propositions.
In such a representation, a domain model which almost interprets all partially observed traces is learned.

%The inputs of learning phrase are plan traces with specific number of objects and domain knowledge only without action models. The outputs of the learning phrase consist action model for all actions which can explain all the traces, and the GNN which can explain the domain model.
%Given plan traces with specific number of objects and domain knowledge only without domain model, the learning phrase is used to learn the domain model of the input instances, which can explain all the input plan traces.
%We model a state as a graph and a planning process as a graph evolution in a recurrent graph neural network block. After learning based on GNN, our GNN block can generate a complete model of graph neural network block explaining the whole domain model. Moreover, it provides a method to vectorize states and actions that differs from traditional ways which may helps to figure out the knowledge hidden in the plan traces.

The vectorization representation learned in the learning module provides a way to learn a heuristic function via an MLP.
Every pair of states orderly occurring in the estimated plan traces make up of a training set for an action selection network.
The action selection network is trained to return the actions executed in the former state in every pair, which are considered as appropriate actions towards the latter state.
%The estimated plan traces with state vectors and action states are considered as the training set for an action selection network.
%Taking two states in
The heuristic function is obtained via computing the distances to the appropriate actions.
%The heuristic network takes two states in the trace as input and is trained to output an action executed in the former state, which is considered as a suitable action towards the latter state.
Then, the heuristic function learned helps to choose a suitable action towards the goal state in the current state during planning.%when searching for a plan.

%The detached representations of states and propositions allow representing a new state which never occurs in the training set.
%It further allows solving new planning instances in the same domain.
%With the assistance of the heuristic function, a plan for new planning instances will be found if it exists.

%The output in learning phrase turns into the input of the planning phrase. We encode an initial state in the GNN block, and pick an action used by the GNN block which will be applied in the current graph.
%Then the GNN evolves the graph and acquires a new state. %setting specific number of actions
%
%In our framework, an improvement of action selection is being considered. The heuristic function network takes place of random selection of actions. It takes the state vectors as input and acquires an action vector. With the distanced-based approach, the most similar action in the vector space will be picked to be applied in the current state. Then Our GNN-block can play a part of planner. With the assistance of heuristic function, it can work on the new planning problems.

%dataflow
%\vspace{-2mm}
\section{Domain-model learning}
%\vspace{-2mm}
In this section, we propose a sequence-to-sequence domain-model learning framework based on GNN \citep{DBLP:journals/corr/abs-1806-01261} to handle plan traces, which are in the form of sequence.
%show how to learn a domain model based on GNN with a recurrent configuration.
%The input plan traces are sequences, which suggests us to use a recurrent configuration to handle them.

To avoid ambiguity, we use the bold type for vectors: $\vp_j,\vs,\va$ stand for the vectors of the proposition $p_j$, the state $s$ and the action $a$, respectively, and $\mathbf{P}$ stands for the set of proposition vectors.
For a unified representation, we consider they all are $k$-dimension real-number vectors.

%\subsection{Proposition-state Graph}
%To learn a detached representation of propositions and states, we put a state and all propositions as vertexes into a graph.
%Then
%GNN provides an interpretable
We define a proposition-state graph is a tuple $G = (V,E,\va)$ where $V = \{s\} \cup \Prop$ is a set of vertexes
and $E=\{e_j = (p_j,s) | p_j \in \Prop\}$ is a set of directed edges from every proposition vertex to the state vertex, and $\va$ is an action vector with the meaning that action $a$ will be executed in state $s$.
Every vertex is equipped with an attribute of $k$ real numbers, which is considered as their vectorization representation.
Every edge $e_j$ has a boolean attribute $\ve_j$, which captures the interpretation of $p_j$ in $s$, which has a boolean attribute.
Formally, $\ve_j = 1$ if $p_j \in s$; otherwise, $\ve_j=0$.

\subsection{Updating Proposition-state Graphs}
In the start of the learning phase, the proposition vectors and the action vectors are first initialized randomly from uniform distributions.
For every partially observed plan trace, we initialize the proposition-state graph $G^0$ by assigning the edge attributes according to the initial state $s_0$.
% the edge attributes are determined via translating the initial state, which is a set of propositions, by the encoder function $\phi^\mathrm{en}$.
%With a function $\phi^s$

As every state can be represented by propositions and their interpretations,
in a proposition-state graph, the unique state vector is obtained via the vectors of the propositions and the edge attributes.
Formally, for a state $s$, we use a state update function to get its vector $\vs$:
\begin{equation}\vs = \phi^\mathrm{s}(\mathbf{P},\mathbf{E})
\label{fml:vs}
\end{equation}
where $\mathbf{E}$ is the set of the edge attributes.
Obviously, the state vector $\vs$ is determined by the proposition vectors and their interpretation in the state, resulting in that a state uniquely corresponds to a state vector.
%The vector $\vs$ represents a state when the interpretation of every propositions are known.
To learn the state update function $\phi^\mathrm{s}$, we use an MLP which takes the concatenation of the proposition vectors and the edge attributes as input.

%In the encoder phrase, the initial state is also need to be encoded. We finish this by translating the relationship between the state and the propositions into the edges' attribute vector. If the proposition $p_k$ is in the state, the value of the output edge $\boldsymbol{e_k}$ is 1, otherwise the value is 0.
%
%The proposition vectors first initialize randomly. But the vector of the state node is required to be consistent in the planning phrase, so it can not be random. And considering the characteristics of the graph structure and planning problem area, it is designed to be depend on the proposition-state edges and  proposition vectors.
%\begin{center}
%\begin{equation}
%v(s) = \phi^s (E, v(P)) % 怎么表达concreate
%\end{equation}
%\end{center}
%We implement a function that takes the proposition-state edges' attribute vectors and proposition vectors as input and get the state vector as output by training.

%\begin{figure}[!htp]
%\centering
%\includegraphics[width=\linewidth]{GNN.png}
%\caption{The Recurrent Framework for domain-model learning}
%\label{fig:recurrent}
%\end{figure}

%\subsection{state-transition}
To formalize the progression of a state caused by an action execution, we define that a proposition-state graph $G=(V,E,\va)$ is updated by applying the action vector $\va$. %proposition-state graph $G'=(V',E',\va')$.
%Taking advantages of the configurableness feature of GNN models, we construct a recurrent framework which consists of state-transition blocks to capture the plan traces (as shown in Figure \ref{fig:recurrent}).
In our sequence-to-sequence framework, the sequential actions in the input plan trace lead that the proposition-state graph is updated continuously.

%The state-transition block takes a proposition-state graph $G=(V,E,\va)$ as input and returns a updated proposition-state graph $G'=(V',E',\va')$.
%The state-transition block  consists of three functions: the edge update function $\phi^\mathrm{E}$, the state update function $\phi^\mathrm{s}$ and the decoder function $\phi^\mathrm{de}$.
In the proposition-state graph, the action vector first changes the edge attributes.
%
%The update of the proposition-state graph starts with the replacement of the action vector.
%Then the action vector leads the change of the edge attributes.
Formally, the edge $e_j$ from $p_j$ to $s$, we use an edge update function to obtain its estimated probability $\hat{\ve}_j$:
\begin{equation}\hat{\ve}'_j = \phi^\mathrm{E}(\ve_j,\vs,\vp_j,\va).\end{equation}
By concatenating all the edges, we generalize the edge update function into the edge set:
\begin{equation}\hat{\vE}' = \phi^\mathrm{E}(\vE,\vs,\vP,\va).
\label{fml:vE}
\end{equation}
%Note that
Similarly, to learn the function $ \phi^\mathrm{E}$, we use an MLP which ends with a sigmoid function and
outputs an estimated probability $\hat{\ve}_j$ for every edge attribute $\ve_j$.
To keep the consistency with the interpretation of propositions, the estimated probability needs to be decoded to the boolean edge attribute $\vE'$ by the decoder function $\phi^\mathrm{de}$.

%\vspace{-2mm}
\subsection{Learning Applicable and State-transition Functions}
The change of the edge attributes directly cause the change of the state vector via the state update function $\phi^\mathrm{s}$.
% the vector of the next state $\vs'$ is determined by the current state vector $\vs$ and the action vector $\va$.
Then we define a state-transition function $f$ for the state vector and the action vector:
\begin{align}
  \vs' & = \phi^\mathrm{s}(\vP,\vE') \nonumber \\
   &=   \phi^\mathrm{s}(\vP,\phi^\mathrm{de}(\phi^\mathrm{E}(\vE,\vs,\vP,\va)))\label{fml:state-trans}\\
   &= f(\vs,\va) \nonumber
\end{align}
%As the edge attribute set $\vE$ depends on the state $s$, according to equations (\ref{fml:vs}) and (\ref{fml:vE}), once all proposition vectors in $\vP$ are learned, the state-transition function $f$ is deterministic.
According to equations (\ref{fml:vs}) and (\ref{fml:vE}), once all proposition vectors are learned, then $\vP$ will not change and $\vE$ are determined by the state vector $\vs$ and the action vector $\va$.
In other words, %the state-transition function $f$ is deterministic and
the next state $s'$ is uniquely determined by the current state $s$ and the action $a$ executed.

A partially observed plan trace yields a sequence of proposition-state graphs by replacing the actions. %and every edge attribute set in the sequences is the estimated state.
Formally, for a plan trace $\hat{\pi}= \langle s_0,a_1,...,s_n \rangle$, we use $G^i=(V^i,E^i,\va_{i+1})$ for $0 \leq i < n$, %where $E^i=\{e^i_1,...e^i_m\}$,
to denote the $i$th proposition-state graphs in the corresponding sequence.
Then every edge attribute set $\vE^i$ stands for an estimated state $s_i$ in the sequence.
To train the functions $\phi^\mathrm{s},\phi^\mathrm{E}$ and the vectors of propositions and actions correctly, we define a loss function to evaluate the differences between the estimated states and the observed states.
The estimation on the propositions in a state can be considered as a logistic regression problem for the observed propositions in the state, which suggests us to employ the cross-entropy loss function. %: %For that, we use the cross entropy loss as our loss function.
%%Given a partially observed trace $s_0,a_1,...,s_n$, we use $G^i=(V^i,E^i,\va_{i+1})$ for $0 \leq i < n$, %where $E^i=\{e^i_1,...e^i_m\}$,
%%to denote the $i$th proposition-state graphs in the corresponding sequence.
%%Then, every edge attribute set $\vE^i = \{\ve^i_1,...\ve^i_m\}$ in the sequence is the estimation of the state $s_i$ in the trace.
%%To evaluate the predict probability of edge attributes between the observations,
%%we employ the cross-entropy loss as the objective function of the recurrent framework as follows:
%%$$L = -\sum_{i=0}^{n} \sum_{p_j \in \obs(s_i)} (\vp^i_j \log \ve^i_j + (1-\vp^i_j)\log (1-\ve^i_j)).$$
%\begin{equation}L = - \frac{1}{n}\sum_{i=0}^{n} \sum_{p_j \in \obs_i} (\vp^i_j \log \hat{\ve}^i_j + (1-\vp^i_j) \log (1-\hat{\ve}^i_j) )\end{equation}
%%In the loss function, only the observed propositions are considered and if the estimated interpretation of a proposition contradicts with its observation, it generates a loss.
%where $\hat{\ve}^i_j$ is the estimated probability for the edge $e^i_j$ in $E^i$.

After prorogating the gradient descent to the functions and the vectors of propositions and actions, when the loss function converges, the state-transition function  is learned.

To learn the applicable function $\alpha$,
for every action $a$,
we consider the intersection of the estimated states in which action $a$ is executed as the precondition of action $a$, denoted by $\pre(a)$.
%The those actions never occurring in the
Then for every state $s$, we define its applicable action set as $\alpha(s) = \{a | \pre(a) \subseteq s\}$.
From a safety perspective, the actions never occurring in the input plan traces are not considered as applicable in any state.
When the model converges, the functions $\phi^\mathrm{s},\phi^\mathrm{E},\phi^\mathrm{de}$ and the proposition vectors $\mathbf{P}$ and all action vectors are learned and fixed.
We can bridge every state and its vector uniquely, \emph{i.e.}, $\vs = g(s)$.
%Then for any state which consists of the propositions in $\Prop$, we define a function $g$ to link the state and its vector:
%\begin{equation}\vs = g(s) = \phi^\mathrm{s}(\mathbf{P},\phi^\mathrm{en}(s)).\end{equation}
%Obviously, the function $g$ is deterministic and every state $s$ have a unique vector $\vs$.
Based on the embedding of propositions, states, and actions, we
represent a planning problem as a tuple $\planP= (\vS,\vs_0,\vS_g,\vactions,\alpha,f)$
where $\vS$ is a set of state vectors, $\vs_0$ is the initial state vector, $\vS_g \subseteq \vS$, $\vactions$ is a set of action vectors and $\alpha, f$ are the applicable and state-transition function, respectively.

%Indeed, the declarative planning language STRIPS provides a compact representation for classical planning.
%A planning problem in STRIPS is a tuple $\stripsP = (\Prop,I,G,\operators)$ where $\Prop$ is a set of propositions, $I$ is the set of propositions that are true initially, $G$ is the set of propositions that are intended to be true together, and $\operators$ is the set of operators where every operator is given as a tuple of proposition sets, $o = (\pre(o),\add(o),\del(o))$, with meaning its precondition, positive effect and negative effect.
%Actually, despite the $\poss$ function, a planning problem in STRIPS also be translated into our vectorization representation as follows: $S = \{g(s)| s \subseteq \Prop\}$, $s_0 = g(I)$, $S_g = \{g(s_g) | G \subseteq s_g\}$, $\actions$ is the set of vectors for operators, and the applicable function $\alpha$ and state-transition function $f$ is defined as
%\begin{align}
%\alpha(\vs) &= \{\va | \pre(a) \in s\}\\
%f(\vs,\va) &= g(\gamma(s,a))=g(((s \cup \add(a)) \setminus \del(a)).\end{align}
%\vspace{-2mm}
\section{Planning with Heuristics}
%In this section, we show how to plan based on the state-transition function learned from the learning module.
%\vspace{-2mm}
\subsection{Learning Heuristics Function}
The heuristic function plays an important role in the forward-search planning techniques, which helps the planner to select suitable actions towards the goal state.
A suitable heuristic function may speed up the problem-solving significantly.
With various heuristic functions being proposed, there is not an approach to choose suitable heuristic function automatically for different planning domains.
For that, we propose an approach to learn the heuristic function based on the embedding of states and actions.

%In planning technique based on forward search, the heuristic function of action selection is an extremely important part. During the search on heuristic solution, an agent will estimate the cost needed from a state to the goal state according to the heuristic function, it can select the action which costs least to execute. By means of planning with heuristic function, we are able to avoid traversing the whole search spaces, which helps a quick planning.

%Our framework can vectorize the states and actions. In addition, it proposes an approach towards generating heuristic function of multi-dimension vector. By vectorizing states and actions, we are able to acquire the general meaning in specific domain after learning from a large training set.
%\begin{definition}

%Given a plan trace $s_0,a_1,s_1,...,s_n$ with a total observation, the same state may occur in multiple positions and
%we use $\pos(s)$ to denote the set of the positions where the state $s$ occurs in the trace.
%Then for a set of totally observed traces,
%we define the action selection function $g: S \times S \longrightarrow \actions$ such that
%for any two states $s,s'$, $h(s,s')= a_{i{+}1}$ where $\gamma(s_i,a_{i{+}1})$ and $(i,j)$ is the pair with the closest distance between the positions in $\pos(s)$ and $\pos(s')$, with $i < j$.
%Intuitively, the action selection function outputs the action executed in the former state in the trace, which makes the state progress towards the latter state.
%In particular, if the pair of states have multiple occurrences, we consider the closest pair.

Given a set of fully observed plan traces $\pi_k = \langle s_0,a_1,s_1,...,s_n \rangle$ %with the total observation,
we define the action selection function $h: S \times S \longrightarrow 2^\actions$ such that $a_{i{+}1} \in h(s_i,s_j)$ where $i<j$ and $a_{i{+}1}$ is the action executed in the state $s_i$ in some $\pi_k$.
As the same state pair may occur in different traces, there are more than one actions executed in the former state $s_i$.

With the embedding of states, we generate tuples $(\vs,\vs',a)$ where $a \in h(s,s')$ as training examples from the estimated trace set obtained from the learning module.
%to learn the action selection function for the vectorization representation.
%But in fact the state pairs do not often occur in the trace set,
%we keep only one training tuple $(\vs,\vs',\va)$ for each state pair $(s,s')$ according to
Actually, it is a multi-label learning task \citep{zhang2014review}.
Then we construct an MLP $\phi^\mathrm{h}$ which takes the concatenation of two state vectors as input and outputs a list of recommendation confidences for every action.
We train the network to minimize the sigmoid cross-entropy loss between the recommendation confidences and the action labels $h(s,s')$.
%To learn an action selection function for state vectors, we construct a neural network $\phi^\mathrm{h}$ which takes the concatenation of two state vectors as input.
%The neural network $\phi^\mathrm{h}$ ends with a sigmoid function and outputs an estimated probability vector in the one-hot representation.
%Suppose the probability vector $\hat{\mathbf{v}}=[\hat{v}_1 \ ...\ \hat{v}_{|\actions|}]$ is the output of $\phi^\mathrm{h}(\vs,\vs')$,
%we evaluate the difference between the prediction $\hat{\mathbf{v}}$ and the training tuple $(\vs,\vs',\mathbf{v})$
%with $\mathbf{v}=\{v_1,...,v_{|\actions|}\}$,
%by employing a cross entropy loss function:
%\begin{equation}L_h = - \frac{1}{|\actions|} \sum_{i} (v_i \log \hat{v}_i + (1-v_i) \log (1-\hat{v}_i))\end{equation}
%%max( \mathbf{0}, \mathbf{1} +\phi^\mathrm{h}(\vs,\vs') - \mathbf{v})$$
%where $|\actions|$ is the number of actions.
%%$h(s_i,s_j)=a_{i+1}$ for $0 \leq i <j \leq n$, and in the case that $s_i = s_{i'}$ and $s_j = s_{j'}$,
%%$h()$
%%
%%Given a set of  $\Pi$, action vector selection function is defined as:$g: \mathcal{R}^k \times \mathcal{R}^k \rightarrow \mathcal{R}^k$. For all state sequence $s_0, s_1, ..., s_n \in \Pi$, the action vector selection function is required to satisfy $h(v(s_k), v(s_n)) \approx v(a_k)(k = 0, 1, ..., n-1)$, where action $a_1$ is the action which will be applied on state $s_1$.
%%%\end{definition}

%Intuitively, the action selection function provides a set of recommend actions to lead towards the latter state.
Consider the latter state as the goal state,
the action selection function provides a set of recommended actions to lead towards the goal state.
For the current state $s_i$ and the goal state $s_g$, we define a goal-driven heuristic function for every action as its recommendation confidence output by $\phi^\mathrm{h}(\vs_i,\vs_g)$.
%The heuristic function indicates the suitableness probability for every action and the action with the highest value will be chosen to execute.
%
%It can also make full use of the hidden information in the state sequence to find the action leading to the goal state. And according to action vector selection function, the heuristic function of the action is designed as:
%\begin{definition}
%Given a state $s_i$ and goal state $s_g$, the heuristic function of selecting the next action is:
%\begin{center}
%    $h(a_{i+1} = \vert h(v(s_i), v(s_g)) - v(a_{i+1} \vert$
%\end{center}
%\end{definition}
%The action with the lowest heuristic value is what we need to choose, which is implied to be the most possible correct action according to the training set.

%Our framework proposes a planning technique based on graph neural network and action selection heuristic function, which greedily selects an action with lowest heuristic value all the time and iteratively updates the state successor function to get a new state until arrives the goal state.

%\vspace{-2mm}
\subsection{Planning with Heuristics Learned}
%Next, we show how to solve the new planning instances under the domain model learned and the heuristic function learned.
%\vspace{-2mm}
%For a new planning instance $(s_0,s_g)$, we first encode the two states into state vectors $\vs_0 = g(s_0)$ and $\vs_g = g(s_0)$. %via the state update function $\phi^\mathrm{s}$ and the proposition vectors $\mathbf{P}$.

Based on the learned domain model,
%recurrent framework of the GNN model,
we propose a progression-based algorithm to compute a plan for the planning instance $(s_0,S_g)$, as shown in Algorithm \ref{alg:Framwork}.
To implement the backtracking, we use a list $Visited$ to record the visited state with the action executed in it and use a list $History$ to record the visited state with the plan executed until it.
We first set the current state as the initial state $\vs_0$ and initialize the two lists to be empty.
Then we start to find a plan via selecting a goal $\vs_g$ state from the goal state set $\vS_g$.
By selecting actions to execute repeatedly, once it reaches one of goal states, the algorithm finds a plan (line 11-12).
%selecting an action to execute repeatedly.
Observe that the action selection function outputs an action set with at most three actions, at every step we choose one of the top 3 recommended actions which are also applicable in the current state (line 5).
Formally, we use $\Phi_3^{\vs,\vs_g}$ to denote the set of actions with the three highest recommendation confidences in $\phi^\mathrm{h}(\vs,\vs_g)$.
Once an action is executed, we update the current state, the current plan and the two lists $Visited$ and $History$ (line 7-12).
When all applicable actions in $\Phi_3^{\vs,\vs_g}$ have been visited, the algorithm have to backtrack to the last state $s^-$ via a \textsc{Pop}~ operator on $History$ (line 15-16).
The current plan $\sigma$ should be regressed by removing its last action, which is done via a \textsc{Regress}~ operator (line 17).
Once the list $History$ becomes empty again, it means every possible recommended action sequence cannot achieve the selected goal state and it needs to choose another goal state (line 13-14).
When every goal state are tried and no plan is found, the algorithm returns failure.
%%Furthermore, the action which have been executed in the same state is forbidden (line 6).
%
%%Every time we choose the action which has the highest value in the heuristic function $\phi^\mathrm{h}$ for the current state towards the goal state (line 5).
%Next, the action selected influences the state via the deterministic state-transition function $f$, which leads to the vector of the next state.
%According to equation (\ref{fml:state-trans}), the state-transition function $f$ first updates the proposition-state graph corresponding to the current state and the action selected
%and outputs a new proposition-state graph with different edge attributes, and then updates the state attribute vector via the function $\phi^\mathrm{s}$.
%Then we update the current state and the current plan (line 7-8).
%After repeating the process, it terminates when the goal state is achieved (line 9-10).
%
%In the first place, the initial state and goal state are encoded by the encoder. Then we select an action, applying it on the graph and adding it into the plan $\pi$. Since the actions that  have learned in the learning phrase store in the GN block, we can update the edges according to equation(2). Making use of the edges from decoder, the state node follows the update. After updating edges and state nodes, the graph evolves as the state steps forward. If the state has arrived the goal state, the planning will be finished.
%
%\vspace{-2mm}
\begin{algorithm}[!htp]
  \caption{$\textsc{GNN-Plan}\ (\planP, \phi^\mathrm{h})$}
    \label{alg:Framwork}
%  \begin{algorithmic}[1]
    \Input{ A planning problem $\planP=(\vs_0,\vS_g,\vA,\alpha,f)$, an action selection network $\phi^\mathrm{h}$ %and an empty plan $\epsilon$
%      the initial state $s_0$,
%      the goal state $s_g$, the action set $\actions$,
%      the state-transition function $f$,
%      the heuristic function $\phi^\mathrm{h}$
      }
    \Output
      {A plan $\sigma$}
  %  \Function {$\textsc{GNN-Plan}\ $} {$f_p$, $f_h$, $s_0$, $s_g$, $Act$}
  %$\pi \leftarrow \epsilon$;\\
 % \uIf {\vs_0 \in \vS_g} {\textbf{return} $\epsilon$};\\
         $\vs \leftarrow \vs_0$;\\
         $Visited \leftarrow History \leftarrow \emptyset$;\\
         %$trace \leftarrow$ $(\vs,\pi)$;\\
 \For {each $\vs_g \in \vS_g$}{
%        $v(s_g) = f_p(s_g)$;\\
%        $v(s) = f_p(s_0)$;\\
    \While {true} {
        \For{each $a \in \Phi_3^{\vs,\vs_g} \cap \alpha(s)$}{
            % $v \leftarrow f_h(v(s), v(s_g))$
            % \State
            %pick $a$ in $\phi^\mathrm{h}(\vs,\vs_g)$ with the highest value;\\
            % \State
            %$a = \textsc{Sel}(s, s_g)$;\\
            \uIf {$(\vs,a) \not\in Visited$}
                {$Visited \leftarrow Visited \cup (\vs,a)$;\\
                 $\sigma \leftarrow \sigma \circ a$;\\
                 $History \leftarrow History \cup (\vs,\sigma)$;\\
                 $\vs \leftarrow f(\vs,\va)$;\\
                \uIf { $\vs \in \vS_g$ } {\textbf{return} $\sigma$;}
            }
        }%for2
        \uIf {$History == \emptyset$} {\textbf{break};}
        $(\vs^-,\sigma) \leftarrow \textsc{Pop}(History)$;\\
        $\vs \leftarrow \vs^-$;\\
        $\sigma \leftarrow \textsc{Regress}(\sigma)$;\\
     }%while
  }%for1
  \textbf{return} fail;\\
 %\end{algorithmic}
 %%\vspace{-2mm}
\end{algorithm}

%\vspace{-3mm}

\section{Experiment}
%\vspace{-2mm}
We apply \solver on five classical planning domains including Logistics, ZenoTravel, Depots, Ferry and Mprime, and compare \solver to the classical domain-model learning system ARMS\citep{DBLP:journals/ai/YangWJ07} which invokes a MAX-SAT solver. \solver is implemented in Tensorflow and GNN framework\footnote{\url{https://github.com/deepmind/graph_nets}}, and it takes approximately three hours to train on a single GPU GeForce RTX 2080 Ti\footnote{Due to the size limit on the supplemental materials, the omitted data, code, and supporting materials are available on({\scriptsize\url{https://tinyurl.com/NeurIPS19-231}}).}.

%\vspace{-2mm}
\subsection{Data}
%\vspace{-2mm}
\begin{wraptable}{r}{5.4cm}
%\begin{table}[h]
\caption{Domain Details}
%\vspace{-3mm}
\label{tab:details}
\small
\begin{center}
\begin{tabular}{|r|r|r|}

\hline

           & proposition &     action \\

\hline
 Logistics &        137 &        150 \\
\hline
     Depot &        110 &        115 \\
\hline
Zeno-Travel &        131 &        279 \\
\hline
     Ferry &         99 &        126 \\
\hline
    Mprime &        216 &        791 \\
\hline
\end{tabular}
\end{center}
%\end{table}
%\vspace{-2mm}
\end{wraptable}
%\vspace{-2mm}
We first get the problem generators from FF planner homepage\footnote{http://fai.cs.uni-saarland.de/hoffmann/ff-domains.html}(for Logistics, Ferry, Mprime) and International Planning Competition website\footnote{http://ipc02.icaps-conference.org/} (for Depots, Zeno-Travel). We randomly generate 2100 disjoint planning instances for each domain and take 2000 instances as the training set and 100 instances as the testing set. Table \ref{tab:details} shows the upper bound of the number of propositions and actions in each domain. For the plan traces with fewer propositions, we use a 0-padding method.
By invoking FF planner, we generate a plan for each planning instances and further obtain 2100 plan traces.
To capture the partial observation, we randomly remove the propositions according to the partial observation percentages (0\%, 20\%, 40\%, 60\%, 80\%, 100\%) from every intermediate state in the plan traces.

\subsection{Training Details}
The training phase is divided into two parts in \solver. First, we trained the sequence-to-sequence model to acquire a domain model with the functions $\phi^\mathrm{s},\phi^\mathrm{E},\phi^\mathrm{de}$ and the vectorization representations of propositions and actions.
These functions are designed as a two-layer MLP respectively and each layer has 100 neurons and the layer normalization.
Second, we train an action selection network $\phi^\textrm{h}$ on the estimated plan traces, which is designed as three-layer network with the layer normalization and has 150 neurons at each hidden layer.
%Further than that, we try to replace the MLP with the support vector machine and Random Forest.

The action vectors and proposition vectors are represented as 100-dimensional vector ($k=100$). %while the edge is represented as 1 dimensional vector. And
The action vector and proposition vector are initialized uniformly and randomly within the range [-0.6, 0.6] ~\citep{DBLP:journals/jmlr/GlorotB10} ($\frac{6}{\sqrt{k}} = 0.6$). We train our model using the Adam optimizer~\citep{DBLP:journals/corr/KingmaB14} with a batch in size of 20 and an initial learning rate of  $10^{-3}$.
%. The learning rate of the optimizer is initialized as $10^{-3}$ and other hyper parameters are set as default.
%The MLP that acts as our action selection function is designed as 2 hidden layer and each layer contains 150 neurons.
%To obtain a better training performance, we also apply layer normalization in the above MLPs.

%dropout and regularizers
%The accuracy of the testing set is unseen in the training phrase.

\subsection{Metrics}

\textbf{Learning Performance Metrics}.
%After a certain number of iterations, we get the resulting domains respectively.
%Then we compare them with the domains that the human encode.
%To evaluate the learning performance of our approach \texttt{LP-GNN}, we compare its learned domain model with the original domain model in each domain which is considered true.
The learning performance of our approach is measured with the precision and recall metrics, by comparing the estimated state sequences with the real ones in the testing set.

%Taking the action sequence in the plan trace as input, the proposition-state graph progresses and a sequence of proposition-state graphs is induced. We obtain the states from the sequence of proposition-state graphs.

%%each of which has an edge attribute set.
%%The edge attribute set of the last proposition-state graph actually is the last states.
%The states induced by executing the action sequences from the initial state under \texttt{LP-GNN} are compared with the ones induced by the artificial domain model.
%We utilize the resulting action model to progress the action sequence from the initial state which already acquired from the collected plan traces and obtain the final state. Then we compare it with the all states in the plan trace that we use.

%We use $Err^i_j$ to denote the error count of propositions in the $j$th state for the $i$th plan trace $T^i$. The total number of atoms in one state is marked with $T(state)$.
%Suppose the cardinality of the proposition set is $m$,
%we define the average accuracy for the state sequences as:
%$$stateAccuracy = 1- \frac{1}{N}\sum_{i = 1}^{N}\frac{\sum_{1 \leq j \leq |T^i|}Err^i_j}{m \cdot|T^i|}$$ where N is the total number of plan traces used in the evaluating process.

%To get a better evaluation on the learning performance, we introduce two metric of precision and recall in classification problem in our evaluation process.
Intuitively, precision gives a notion of soundness while recall gives a notion of the completeness of the estimated state sequences.
We use $t_p$ to denote the propositions both in the real and estimated state, $t_n$ to denote the propositions in neither the real state nor the estimated state, $f_p$ to denote the propositions not in the real state but in the estimated state and $f_n$ to denote the propositions in the real state but not in the estimated state.
Then for an estimated state, we compute its precision by $Precision = \frac{t_p}{t_p + f_p}$ and its recall by $Recall = \frac{t_p}{t_p + f_n}$.
To evaluate the estimation performances of the learning approaches on the testing set,
we generalize these two metrics into state sequence sets by computing their average precision and recall for every state in every sequence.

\textbf{Planning Performance Metric}.
%In order to evaluate the ability to solve new problems of our learned action model, we compute the plans on the testing planning instances based on our learned action model.
As we mention before, domain models in the declarative language are sensitive to their accuracy.
Even though the learned domain models interpret the partially observed plan traces perfectly, it is possible that they cannot be used to solve the real planning problems.
It is more important to evaluate the domain-model learning approaches on the ability of solving real problems.
More specially, for the learned domain model, we generate plans under it for the planning instances in the testing set and test whether these plans are solutions to these planning instances under the original domain models.
If so, the testing instance is considered as solved by the learned domain model.
Then we introduce a metric as the percentages of solved instances on all testing instances, \emph{i.e.}, $I = \frac{\#instances\ solved}{\#testing\ instances}$.
%For that, we introduce a metric on the percentage of solved real planning instances in the

%Taking a planning instance $(s_0, s_g)$ in the testing plan traces as input, we use the learned action selection function to choose actions to compute a plan from $s_0$ to $s_g$ in our framework.
%In order to verify the plan computed by the learned domain model,
%we check whether the plan reaches the goal state under the artificial domain model.
%Once the goal state is reached, the plan computed is effective and solves the planning instances.
%%When we execute all the actions, the current state still doesn't satisfy the goal state, then we will consider it is ineffective.
%%Then we define the instance solved rate as the proportion of the instance solved by the domain model, that is,
%%$$I = \frac{\# instance solved}{25}.$$
%Also, we compare our approach with the learning system ARMS on the ability to solve new instances.
%We use the domain model learned by ARMS and invoke FF planner to compute plans.
%Then we calculate the number of instances solved by these plans and define the percentage of instances solved as the proportion of instances solved by the plans computed by the domain model learned among the total number of testing instances, that is,:
%$$I = \frac{\#instances\ solved}{N}$$
%where N is the size of the testing planning instances.

%\vspace{-2mm}
\subsection{Results}
\vspace{-2mm}

%We also compare the average error rate of the last states of our approach with that of the classical action model learning system ARMS \cite{DBLP:journals/ai/YangWJ07}.
%The results of the experiments on the three domains are shown in Figure \ref{fig:experiment1}.

\vspace{-1mm}
\begin{table}[h]
	\caption{Learning Performance in the Five Domains under Various Observation Percentages}
	\begin{center}
\vspace{-1mm}
\resizebox{\linewidth}{30mm}{
\begin{tabular}{|c|c|c|c|c|c|c|c|c|c|c|c|c|}
\toprule
\multirow{3}{*}{Domain} &                        \multicolumn{ 4}{|c|}{0\%} &                       \multicolumn{ 4}{|c|}{20\%} &                       \multicolumn{ 4}{|c|}{40\%} \\
\cline{2-13}
\multicolumn{ 1}{|c|}{} & \multicolumn{ 2}{|c|}{\solver} & \multicolumn{ 2}{|c|}{ARMS} & \multicolumn{ 2}{|c|}{\solver} & \multicolumn{ 2}{|c|}{ARMS} & \multicolumn{ 2}{|c|}{\solver} & \multicolumn{ 2}{|c|}{ARMS} \\
\cline{2-13}
\multicolumn{ 1}{|c|}{} &      P(\%) &      R(\%) &      P(\%) &      R(\%) &      P(\%) &      R(\%) &      P(\%) &      R(\%) &      P(\%) &      R(\%) &      P(\%) &      R(\%) \\
\midrule
 Logistics &      87.09 &      {\bfseries 66.33} &     {\bfseries 88.39} &      46.68 &     {\bfseries 99.37} &      98.64 &      90.21 &      {\bfseries  100} &      {\bfseries 99.53} &      99.34 &      90.21 &      {\bfseries  100} \\
\hline
Zeno-Travel &      82.71 &       61.3 &     {\bfseries 95.19} &     {\bfseries 68.08} &      99.11 &     {\bfseries 98.69} &        {\bfseries 100} &      95.96 &      99.77 &      {\bfseries 99.75} &     {\bfseries   100} &      95.96 \\
\hline
     Depot &      83.14 &      {\bfseries 82.08} &     {\bfseries 95.62} &      71.85 &     {\bfseries 98.65} &      {\bfseries99.74} &      93.46 &      92.93 &      {\bfseries 98.49} &     {\bfseries 99.88} &      93.46 &      92.93 \\
\hline
    Mprime &      91.21 &     {\bfseries 67.13} &      {\bfseries97.09} &      61.85 &     {\bfseries 92.62} &      88.84 &      91.29 &     {\bfseries 99.91} &     {\bfseries 95.21} &      93.38 &       90.3 &     {\bfseries 99.91} \\
\hline
     Ferry &     {\bfseries 96.91} &      {\bfseries 79.51} &      96.42 &      66.43 &      {\bfseries99.98} &      99.81 &      98.58 &       {\bfseries 100} &        {\bfseries 100} &      99.84 &      98.58 &     {\bfseries   100} \\
\toprule
\multirow{3}{*}{Domain} &                       \multicolumn{ 4}{|c|}{60\%} &                       \multicolumn{ 4}{|c|}{80\%} &                      \multicolumn{ 4}{|c|}{100\%} \\
\cline{2-13}
\multicolumn{ 1}{|c|}{} & \multicolumn{ 2}{|c|}{\solver} & \multicolumn{ 2}{|c|}{ARMS} & \multicolumn{ 2}{|c|}{\solver} & \multicolumn{ 2}{|c|}{ARMS} & \multicolumn{ 2}{|c|}{\solver} & \multicolumn{ 2}{|c|}{ARMS} \\
\cline{2-13}
\multicolumn{ 1}{|c|}{} &      P(\%) &      R(\%) &      P(\%) &      R(\%) &      P(\%) &      R(\%) &      P(\%) &      R(\%) &      P(\%) &      R(\%) &      P(\%) &      R(\%) \\
\hline
 Logistics &      {\bfseries 99.88} &      98.96 &      90.21 &     {\bfseries   100} &      {\bfseries99.85} &       99.5 &      90.21 &       {\bfseries 100} &      99.99 &      99.83 &      {\bfseries  100} &      {\bfseries  100} \\
\hline
Zeno-Travel &      99.85 &     {\bfseries 99.79} &      {\bfseries  100} &      95.96 &       99.9 &      {\bfseries 99.9} &       {\bfseries 100} &      95.96 &      99.83 &     {\bfseries 99.79} &      {\bfseries  100} &      95.96 \\
\hline
     Depot &      {\bfseries 98.64} &      {\bfseries 99.96} &      93.46 &      92.93 &      {\bfseries 98.21} &     {\bfseries 99.89} &      93.46 &      92.93 &     {\bfseries  99.97} &      {\bfseries 99.97} &      93.46 &      92.93 \\
\hline
    Mprime &     {\bfseries 97.5}1 &      96.08 &      91.29 &     {\bfseries 99.91} &     {\bfseries 98.16} &      95.89 &      91.29 &     {\bfseries 99.91} &     {\bfseries 98.29} &      96.49 &      97.58 &    {\bfseries  99.98} \\
\hline
     Ferry &        100 &     {\bfseries 99.84} &        100 &      99.06 &       {\bfseries 100} &      99.83 &      99.22 &       {\bfseries 100} &        100 &        100 &        100 &        100 \\
\hline
\end{tabular}
}
	\end{center}
{\small P = the average precision, R = the average recall, 0\%, 20\%, 40\%, 60\%, 80\%, 100\% are observation percentages.}
%%\vspace{-2mm}
\end{table}
\vspace{-2mm}

Table 2 shows the learning performances of our approach \solver and ARMS on the testing set.
With the observation percentage increasing, both the approaches have better and better performances on estimating states.
The results show that \solver and ARMS are comparable on the learning performance.
In \solver, the loss on the training set are various observation percentages less than $10^{-5}$, which means that the learned domain models almost interpret all training plan traces and can be considered as solutions to the learning problems.
While in ARMS, all plan traces are interpreted and it is because it is based on a MAX-SAT solver.

\begin{figure*}[htpb]
%	%\vspace*{3mm}
	\centering
	\subfigure[Depots]{
		\includegraphics[width=0.28 \textwidth,clip]{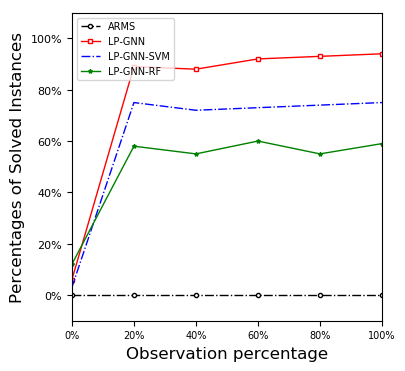}
	}
	\subfigure[Ferry]{
		\includegraphics[width=0.28 \textwidth,clip]{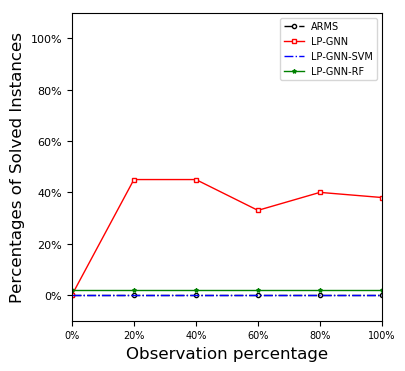}
	}
	\subfigure[Logistics]{
		\includegraphics[width=0.28 \textwidth,clip]{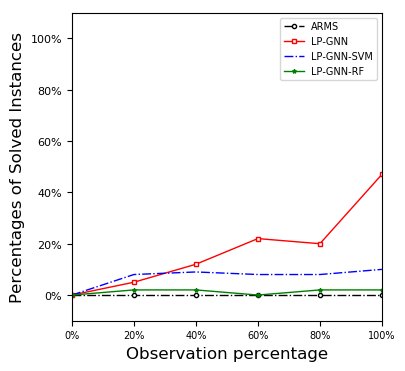}
	}
	\subfigure[Mprime]{
	\includegraphics[width=0.28 \textwidth,clip]{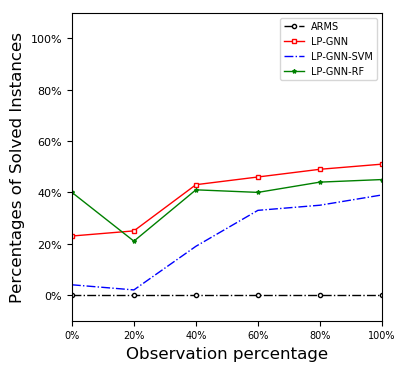}
	}
	\subfigure[Zeno-Travel]{
	\includegraphics[width=0.28 \textwidth,clip]{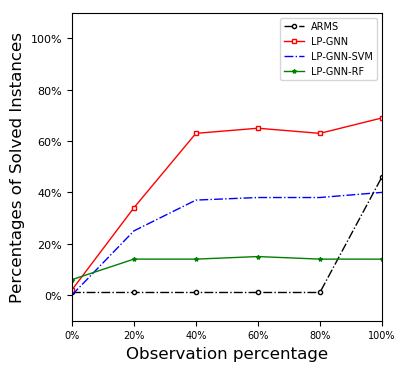}
	}
	\caption{{\small Comparisons on instances solved with various observation percentages. \solver is our approach and \solver-SVM and \solver-RF are our approaches with replacing action selection MLP by SVM and Random Forest. Instances solved are the testing instances which are solved under the original domain model by the plans computed according to the learned domain model.}}
	\label{fig:experiment1}
%%\vspace{-2mm}
\end{figure*}
%%\vspace{-2mm}

%According to the Table 2., it is obvious that our model performs well in state complement.

To evaluate the real problem solving ability, we compare our approach \solver against ARMS on the percentages of instance solved on the testing set, whose experimental results are shown in Figure \ref{fig:experiment1}.
As ARMS outputs domain models in STRIPS, we call FF planer to generate plans.
Obviously, our approach significantly outperforms ARMS on the ability of solving real problems and the domain model learned by ARMS fails to solve any real problems except for the Zeno-Travel domain.

For the model ablation, in \solver we replace the action selection MLP by SVM and Random Forest, and modify the \textsc{GNN-Plan} planning algorithm accordingly.
We evaluate the effectiveness, on the real planning problems, of plans computed by these three algorithms under the same learned domain models.
The results show that learning action selection policy via MLP outperforms other two approaches. %is the most constructive
Actually, for the solved instances, the \textsc{GNN-Plan} planning algorithm with MLP almost generates plans identical with the plans generated by the original domain model using FF planner, which shows the excellent ability on guiding planning of our heuristics learning approach.
%We

%From Figure \ref{fig:experiment1}, we can clearly see that the percentage of instances solved of our approach \texttt{LP-GNN} is higher than that of ARMS.
%Actually, the plan generated under the domain model learned by ARMS hardly solves the testing planning instances.
%%The main reason lies in that some effects of actions are not learned correctly.
%
%
%On the other hand, our approach has a stronger ability to solve new instances.
%Although our approach solves more instances, the plans computed by our approach have many redundant actions because of the heuristic function learned.
%As the precondition of action is not considered in our approach, we drop the action whose precondition is not satisfied under the actual domain model in the plan computed and execute the next action in the plan computed.
%Despite the different partial observation percentage, once the instances whose goal state is learned correctly and which at least have an effective plan under the actual domain model, we can find an effective plan to solve them under the domain model learned by \texttt{LP-GNN}.
%Even though the accuracy of \texttt{LP-GNN} is lower than that of ARMS, our approach has a higher fault tolerance and is able to solve more new instances.
\vspace{-3mm}
\subsection{Analysis}
\vspace{-3mm}
The reason why the learned domain models by ARMS hardly solve any real problems should be rooted in the fact that the plan-searching is extremely sensitive to its accuracy in the declarative language.
From the experimental results, we observe that in Logistics domain, the proposition `(airport ?location)' is learned as an effect of the action `(load-truck ?object ?truck ?location)' in the domain model learned by ARMS. Once the action is executed, the city center where the package is loaded into the truck becomes an airport, resulting in that the airplane can fly to the city center. So, the plan including the action that the plane flies to the city center is generated via the FF planner, but it is not allowed in the artificial domain model. Then, mostly the plans generated under such a learned domain model are ineffective and no instances are solved.
%It demonstrates that the actual problem-solving is very sensitive to the accuracy of the domain model learned by the learning approaches for STRIPS or PDDL language.

The failure of the plans generated by \solver on some instances is blamed for the precondition learned.
Because actions are not sufficiently occurred in the plan traces, if we consider the intersection of the false propositions into the action precondition, it will make the precondition too strong to be satisfied by other states. So, we only focus on the true propositions which, on the other hand, is too weak so that the planning algorithm may execute a unapplicable action.

\section{Related Work}
Domain-model learning has been obtained a lot of attention and there exist a number of approaches \citep{DBLP:journals/ker/AroraFPMP18}.
In this paper, we focus on the learning approaches which return domain models in a declarative language, such as PDDL and its fragments. %$, which are summarized in Table \ref{tab:comparing}.
LOCM~\citep{DBLP:journals/ker/CresswellMW13} and its successor LOCM2~\citep{DBLP:conf/aips/CresswellG11} learn the object-centered representation based on a set of parameterized finite state machines. But these two approaches only can learn action effects on dynamic predicates and fail to handle static predicates which do not change due to action executions.
NLOCM~\citep{DBLP:conf/aips/GregoryL16} extends finite state machines with numeric weights to learn action costs.
PELA~\citep{DBLP:conf/aips/0004ATRI16} refines the input domain model based on top-down induction of decision trees but assume the input domain model to be correct.
OBSERVER~\citep{DBLP:conf/aaai/Wang94} is an incrementally learning system which refines the learned domain model by observing the execution traces for the sampled problems. Whereas, its performance is sensitive to the sampled problems and it may suffer from the incomplete or incorrect domain knowledge.
LAMP~\citep{DBLP:journals/ai/ZhuoYHL10} is a framework to learn more complex domain models with quantifiers and logical implication.
\citep{DBLP:conf/aips/AinetoJO18} proposes an approach to compile the learning problem into a classical planning problem, which may suffer from a scale issue.

Another related work is \citep{DBLP:conf/ecai/MouraoPS10}, which considers action-effect learning problems as classifier problems and proposes a learning approach based on a bank of kernel perceptrons. But it only learns action effects and needs a good number of training examples for good performance.

Some approaches require a fully observed environment where we consider a partially observed one.
LOPE~\citep{DBLP:journals/jirs/Garcia-MartinezB00} learns domain models in STRIPS by repeatedly executing actions based on reinforcement learning.
\citep{DBLP:conf/ijcai/SternJ17} provides a safe domain-model learning approach which guarantees the output domain model to generate safe plans.

There are learning approaches taking noisy plan traces as input which suppose that the input actions may be incorrect.
AMAN~\citep{DBLP:conf/ijcai/ZhuoK13} learns domain models from noisy planning traces via probabilistic graphical models and reinforcement learning.
The line of works by Pasula \emph{et al.} [\citeyear{DBLP:conf/aips/PasulaZK04,DBLP:journals/jair/PasulaZK07}] focus on learning STRIPS-like planning rules via adding noisy outcome in their probabilistic model but fail to handle incomplete observations.

As we mention before, ARMS~\citep{DBLP:journals/ai/YangWJ07} is one of the most classical domain-model learning approaches which have inspired a series of learning approaches.
For example, from the perspective of transfer learning, LAWS~\citep{DBLP:conf/aips/ZhuoYPL11} takes other domain models into account and measures the similarity between the source domains and the target domain via web searching.
For another example, Lammas~\citep{DBLP:conf/atal/ZhuoMY11} learns multi-agent domain models by constructing constraints about agent actions and invoking a MAX-SAT solver.
Besides, CAMA~\citep{DBLP:conf/aaai/Zhuo15} integrates intelligence of crowds into action-model acquisition based on a MAX-SAT solver.
Later \citep{DBLP:journals/ai/ZhuoM014} proposed a learning system HTN-Learner to learn hierarchical task network planning domain models based on a weighted MAX-SAT solver.

Other domain-model learning approaches also concentrate on various inputs.
TRAMP~\citep{DBLP:journals/ai/Zhuo014} and tLAMP~\citep{DBLP:conf/pricai/ZhuoYHL08} use the transfer learning technique and require other domains as inputs, as well as LAWS.
LatPlan~\citep{DBLP:conf/aaai/AsaiF18} proposes an approach to learn action models from fully observed images.

\begin{table}[h]
\centering

\caption{Comparing with PDDL Action Models Learning Approaches}
\label{tab:comparing}
%%\vspace{-1mm}
%\resizebox{0.8\linewidth}{35mm}{
\begin{tabular}{|c|l|l|}

\toprule %添加表格头部粗线

Approaches & Input  & Limitations$\backslash$Features\\

\midrule %添加表格中横线

LOCM, LOCM2 & Action sequences& Only handle dynamic predicates \\

\midrule
NLOCM & Action sequences with costs & Can learn action costs\\

\midrule
PELA  & \tabincell{l}{Initial action models and\\ action sequences} & \tabincell{l}{Assumes correct initial action\\ models}  \\

\midrule
OBSERVER & \tabincell{l}{Action sequence and\\ sampled problems} &\tabincell{l}{Sensitive to the sampled \\planning problems }\\

\midrule
LOPE & repeated action executions& Requires FO environment\\

\midrule
\citep{DBLP:conf/ijcai/SternJ17} & FO plan traces & Requires FO plan traces\\

\midrule
LAMP & \multirow{3}{*}{PO plan traces} & learns ADL domain models\\

\cline{1-1}\cline{3-3}
\citep{DBLP:conf/aips/AinetoJO18}& & compiles into a planning problem\\

\cline{1-1}\cline{3-3}
\citep{DBLP:conf/ecai/MouraoPS10}&  & \tabincell{l}{Only learns effects and requires\\many training examples}\\

\midrule
AMAN & Noisy plan traces & No background knowledge needed\\

\midrule
ARMS &plan traces & \multirow{3}{*}{Call a MAX-SAT solver}\\

\cline{1-2}
Lammas &multi-agent plan traces & \\
\cline{1-2}
CAMA &\tabincell{l}{PO plan traces and\\ crowdsourcing data}& \\

\midrule
LAWS,TRAMP,tLAMP & Plan traces and other domains & Use the transfer learning technique\\

\midrule
LatPlan & Action sequences and images & Requires FO images\\

\bottomrule %添加表格底部粗线

\end{tabular}

%}
\flushleft
FO = fully observed; PO = partially observed
\end{table}
%\vspace{-2mm}
\section{Discussion and Conclusion}
%\vspace{-2mm}

%Different from the work of \citep{DBLP:conf/ijcai/SternJ17} who focus on
Similar with the perspective of \citep{DBLP:conf/ijcai/SternJ17} on the safety of the plans generated by the learned domain model,
in this paper we focus on the effectiveness on the real problems of the plans.
It motivates us to find another way to model domain models which is distinct from the classical declarative language.
Indeed, we aim to learn the vectorization representation of actions, states and propositions in GNN, which actually provides an interpretation for state changes caused by action executions.
By embedding propositions and actions in a graph, the latent relationship between them is explored to form a domain-specific heuristics. Its excellent strength on guiding planning has been demonstrated by the experiment results and we believe that it opens a line of future work on learning domain-specific heuristic functions.
%
%By representing propositions, a state and an action in a graph, GNN actually provides an interpretation for state changes caused by action executions.
%
%Different from the declarative compact representation, in this paper we focus on learning the embedding of propositions, states, and actions in the low-dimension vector space.
%By representing states and actions in a unified vector space, we hope that the information about the states and actions occurring frequently in the training set is able to help the ones occurring less frequently, even never, in the training set to obtain suitable representations.
%Furthermore, the operations in low-dimension vector space generally have a feature of high fault-tolerance, which allows us to compute a plan even though the domain model learned is not sufficient.
%
%In our recurrent framework of GNN models, actions are interpreted as translations operating the proposition-state graphs.
%The configurablness feature of GNN allows capturing the whole plan trace in a recurrent framework and the rational inductive bias in GNN contributes to learning a state-transition function effectively and efficiently.

To sum up, we propose a novel approach \solver to learn the domain model based on GNN from a set of partially observed plan traces.
We first learn the vectorization representations of propositions, states, and actions by putting them into a proposition-state graph.
%the state-transition is accomplished via the progression of the graph caused by an action vector.
The representation in the proposition-state graph allows us to denote new states in the domain and further enables us to solve new planning instances.
%Based on the embedding of states, we propose an approach to learn heuristic function to guide action selection towards the goal.
Finally, we propose a more robust planning framework equipped with a domain-specific heuristic function, which is demonstrated to be more effective on solving real planning problems.
%As the precondition of actions in this paper is not considered, it is interesting to capture action precondition in the graphs.
%Another promising way is to consider noisy in plan traces, which are more common in real-world applications.

\bibliographystyle{named}
\bibliography{ref}

\begin{thebibliography}{}

\bibitem[\protect\citeauthoryear{Aineto \bgroup \em et al.\egroup
  }{2018}]{DBLP:conf/aips/AinetoJO18}
Diego Aineto, Sergio Jim{\'{e}}nez, and Eva Onaindia.
\newblock Learning {STRIPS} action models with classical planning.
\newblock In {\em Proceedings of the Twenty-Eighth International Conference on
  Automated Planning and Scheduling, {ICAPS} 2018, Delft, The Netherlands, June
  24-29, 2018.}, pages 399--407, 2018.

\bibitem[\protect\citeauthoryear{Arora \bgroup \em et al.\egroup
  }{2018}]{DBLP:journals/ker/AroraFPMP18}
Ankuj Arora, Humbert Fiorino, Damien Pellier, Marc M{\'{e}}tivier, and Sylvie
  Pesty.
\newblock A review of learning planning action models.
\newblock {\em Knowledge Eng. Review}, 33:1--25, 2018.

\bibitem[\protect\citeauthoryear{Asai and
  Fukunaga}{2018}]{DBLP:conf/aaai/AsaiF18}
Masataro Asai and Alex Fukunaga.
\newblock Classical planning in deep latent space: Bridging the
  subsymbolic-symbolic boundary.
\newblock In {\em Proceedings of the 32nd {AAAI} Conference on Artificial
  Intelligence, {(AAAI-18)}, New Orleans, Louisiana, USA, February 2-7, 2018},
  pages 6094--6101, 2018.

\bibitem[\protect\citeauthoryear{Battaglia \bgroup \em et al.\egroup
  }{2018}]{DBLP:journals/corr/abs-1806-01261}
Peter~W. Battaglia, Jessica~B. Hamrick, Victor Bapst, Alvaro
  Sanchez{-}Gonzalez, Vin{\'{\i}}cius~Flores Zambaldi, Mateusz Malinowski,
  Andrea Tacchetti, David Raposo, Adam Santoro, Ryan Faulkner, {\c{C}}aglar
  G{\"{u}}l{\c{c}}ehre, Francis Song, Andrew~J. Ballard, Justin Gilmer,
  George~E. Dahl, Ashish Vaswani, Kelsey Allen, Charles Nash, Victoria
  Langston, Chris Dyer, Nicolas Heess, Daan Wierstra, Pushmeet Kohli, Matthew
  Botvinick, Oriol Vinyals, Yujia Li, and Razvan Pascanu.
\newblock Relational inductive biases, deep learning, and graph networks.
\newblock {\em CoRR}, abs/1806.01261, 2018.

\bibitem[\protect\citeauthoryear{Bonet and
  Geffner}{2001}]{DBLP:journals/ai/BonetG01}
Blai Bonet and Hector Geffner.
\newblock Planning as heuristic search.
\newblock {\em Artif. Intell.}, 129(1-2):5--33, 2001.

\bibitem[\protect\citeauthoryear{Bordes \bgroup \em et al.\egroup
  }{2013}]{DBLP:conf/nips/BordesUGWY13}
Antoine Bordes, Nicolas Usunier, Alberto Garc{\'{\i}}a{-}Dur{\'{a}}n, Jason
  Weston, and Oksana Yakhnenko.
\newblock Translating embeddings for modeling multi-relational data.
\newblock In {\em Proceedings of the 27th Annual Conference on Neural
  Information Processing Systems {(NIPS-13)}}, pages 2787--2795, 2013.

\bibitem[\protect\citeauthoryear{Cresswell and
  Gregory}{2011}]{DBLP:conf/aips/CresswellG11}
Stephen Cresswell and Peter Gregory.
\newblock Generalised domain model acquisition from action traces.
\newblock In {\em Proceedings of the 21st International Conference on Automated
  Planning and Scheduling, {ICAPS} 2011, Freiburg, Germany June 11-16, 2011},
  2011.

\bibitem[\protect\citeauthoryear{Cresswell \bgroup \em et al.\egroup
  }{2013}]{DBLP:journals/ker/CresswellMW13}
Stephen Cresswell, Thomas~Leo McCluskey, and Margaret~Mary West.
\newblock Acquiring planning domain models using \emph{LOCM}.
\newblock {\em Knowledge Eng. Review}, 28(2):195--213, 2013.

\bibitem[\protect\citeauthoryear{Edelkamp}{2002}]{DBLP:conf/aips/Edelkamp02}
Stefan Edelkamp.
\newblock Symbolic pattern databases in heuristic search planning.
\newblock In {\em Proceedings of the 6th International Conference on Artificial
  Intelligence Planning Systems}, pages 274--283, 2002.

\bibitem[\protect\citeauthoryear{Franc{\`{e}}s \bgroup \em et al.\egroup
  }{2017}]{DBLP:conf/ijcai/FrancesRLG17}
Guillem Franc{\`{e}}s, Miquel Ram{\'{\i}}rez, Nir Lipovetzky, and Hector
  Geffner.
\newblock Purely declarative action descriptions are overrated: Classical
  planning with simulators.
\newblock In {\em Proceedings of the 26th International Joint Conference on
  Artificial Intelligence {(IJCAI-17)}}, pages 4294--4301, 2017.

\bibitem[\protect\citeauthoryear{Garc{\'{\i}}a{-}Mart{\'{\i}}nez and
  Borrajo}{2000}]{DBLP:journals/jirs/Garcia-MartinezB00}
Ram{\'{o}}n Garc{\'{\i}}a{-}Mart{\'{\i}}nez and Daniel Borrajo.
\newblock An integrated approach of learning, planning, and execution.
\newblock {\em Journal of Intelligent and Robotic Systems}, 29(1):47--78, 2000.

\bibitem[\protect\citeauthoryear{Glorot and
  Bengio}{2010}]{DBLP:journals/jmlr/GlorotB10}
Xavier Glorot and Yoshua Bengio.
\newblock Understanding the difficulty of training deep feedforward neural
  networks.
\newblock In {\em Proceedings of the Thirteenth International Conference on
  Artificial Intelligence and Statistics, {AISTATS} 2010, Chia Laguna Resort,
  Sardinia, Italy, May 13-15, 2010}, pages 249--256, 2010.

\bibitem[\protect\citeauthoryear{Gregory and
  Lindsay}{2016}]{DBLP:conf/aips/GregoryL16}
Peter Gregory and Alan Lindsay.
\newblock Domain model acquisition in domains with action costs.
\newblock In {\em Proceedings of the Twenty-Sixth International Conference on
  Automated Planning and Scheduling, {ICAPS} 2016, London, UK, June 12-17,
  2016.}, pages 149--157, 2016.

\bibitem[\protect\citeauthoryear{Hoffmann and
  Nebel}{2001}]{DBLP:journals/jair/HoffmannN01}
J{\"{o}}rg Hoffmann and Bernhard Nebel.
\newblock The {FF} planning system: Fast plan generation through heuristic
  search.
\newblock {\em J. Artif. Intell. Res.}, 14:253--302, 2001.

\bibitem[\protect\citeauthoryear{Kingma and
  Ba}{2015}]{DBLP:journals/corr/KingmaB14}
Diederik~P. Kingma and Jimmy Ba.
\newblock Adam: {A} method for stochastic optimization.
\newblock In {\em 3rd International Conference on Learning Representations,
  {ICLR} 2015, San Diego, CA, USA, May 7-9, 2015, Conference Track
  Proceedings}, 2015.

\bibitem[\protect\citeauthoryear{Mart{\'{\i}}nez \bgroup \em et al.\egroup
  }{2016}]{DBLP:conf/aips/0004ATRI16}
David Mart{\'{\i}}nez, Guillem Aleny{\`{a}}, Carme Torras, Tony Ribeiro, and
  Katsumi Inoue.
\newblock Learning relational dynamics of stochastic domains for planning.
\newblock In {\em Proceedings of the Twenty-Sixth International Conference on
  Automated Planning and Scheduling, {ICAPS} 2016, London, UK, June 12-17,
  2016.}, pages 235--243, 2016.

\bibitem[\protect\citeauthoryear{Mikolov \bgroup \em et al.\egroup
  }{2013}]{DBLP:journals/corr/abs-1301-3781}
Tomas Mikolov, Kai Chen, Greg Corrado, and Jeffrey Dean.
\newblock Efficient estimation of word representations in vector space.
\newblock {\em CoRR}, abs/1301.3781, 2013.

\bibitem[\protect\citeauthoryear{Mour{\~{a}}o \bgroup \em et al.\egroup
  }{2010}]{DBLP:conf/ecai/MouraoPS10}
Kira Mour{\~{a}}o, Ronald P.~A. Petrick, and Mark Steedman.
\newblock Learning action effects in partially observable domains.
\newblock In {\em {ECAI} 2010 - 19th European Conference on Artificial
  Intelligence, Lisbon, Portugal, August 16-20, 2010, Proceedings}, pages
  973--974, 2010.

\bibitem[\protect\citeauthoryear{Pasula \bgroup \em et al.\egroup
  }{2004}]{DBLP:conf/aips/PasulaZK04}
Hanna Pasula, Luke~S. Zettlemoyer, and Leslie~Pack Kaelbling.
\newblock Learning probabilistic relational planning rules.
\newblock In {\em Proceedings of the Fourteenth International Conference on
  Automated Planning and Scheduling {(ICAPS} 2004), June 3-7 2004, Whistler,
  British Columbia, Canada}, pages 73--82, 2004.

\bibitem[\protect\citeauthoryear{Pasula \bgroup \em et al.\egroup
  }{2007}]{DBLP:journals/jair/PasulaZK07}
Hanna~M. Pasula, Luke~S. Zettlemoyer, and Leslie~Pack Kaelbling.
\newblock Learning symbolic models of stochastic domains.
\newblock {\em J. Artif. Intell. Res.}, 29:309--352, 2007.

\bibitem[\protect\citeauthoryear{Stern and
  Juba}{2017}]{DBLP:conf/ijcai/SternJ17}
Roni Stern and Brendan Juba.
\newblock Efficient, safe, and probably approximately complete learning of
  action models.
\newblock In {\em Proceedings of the Twenty-Sixth International Joint
  Conference on Artificial Intelligence, {IJCAI} 2017, Melbourne, Australia,
  August 19-25, 2017}, pages 4405--4411, 2017.

\bibitem[\protect\citeauthoryear{Wang}{1994}]{DBLP:conf/aaai/Wang94}
Xuemei Wang.
\newblock Learning by observation and practice: {A} framework for automatic
  acquisition of planning operators.
\newblock In {\em Proceedings of the 12th National Conference on Artificial
  Intelligence, Seattle, WA, USA, July 31 - August 4, 1994, Volume 2.}, page
  1496, 1994.

\bibitem[\protect\citeauthoryear{Yang \bgroup \em et al.\egroup
  }{2007}]{DBLP:journals/ai/YangWJ07}
Qiang Yang, Kangheng Wu, and Yunfei Jiang.
\newblock Learning action models from plan examples using weighted {MAX-SAT}.
\newblock {\em Artif. Intell.}, 171(2-3):107--143, 2007.

\bibitem[\protect\citeauthoryear{Zhang and Zhou}{2014}]{zhang2014review}
Min-Ling Zhang and Zhi-Hua Zhou.
\newblock A review on multi-label learning algorithms.
\newblock {\em IEEE transactions on knowledge and data engineering},
  26(8):1819--1837, 2014.

\bibitem[\protect\citeauthoryear{Zhuo and
  Kambhampati}{2013}]{DBLP:conf/ijcai/ZhuoK13}
Hankz~Hankui Zhuo and Subbarao Kambhampati.
\newblock Action-model acquisition from noisy plan traces.
\newblock In {\em {IJCAI} 2013, Proceedings of the 23rd International Joint
  Conference on Artificial Intelligence, Beijing, China, August 3-9, 2013},
  pages 2444--2450, 2013.

\bibitem[\protect\citeauthoryear{Zhuo and
  Yang}{2014}]{DBLP:journals/ai/Zhuo014}
Hankz~Hankui Zhuo and Qiang Yang.
\newblock Action-model acquisition for planning via transfer learning.
\newblock {\em Artif. Intell.}, 212:80--103, 2014.

\bibitem[\protect\citeauthoryear{Zhuo \bgroup \em et al.\egroup
  }{2008}]{DBLP:conf/pricai/ZhuoYHL08}
Hankui Zhuo, Qiang Yang, Derek~Hao Hu, and Lei Li.
\newblock Transferring knowledge from another domain for learning action
  models.
\newblock In {\em {PRICAI} 2008: Trends in Artificial Intelligence, 10th
  Pacific Rim International Conference on Artificial Intelligence, Hanoi,
  Vietnam, December 15-19, 2008. Proceedings}, pages 1110--1115, 2008.

\bibitem[\protect\citeauthoryear{Zhuo \bgroup \em et al.\egroup
  }{2010}]{DBLP:journals/ai/ZhuoYHL10}
Hankz~Hankui Zhuo, Qiang Yang, Derek~Hao Hu, and Lei Li.
\newblock Learning complex action models with quantifiers and logical
  implications.
\newblock {\em Artif. Intell.}, 174(18):1540--1569, 2010.

\bibitem[\protect\citeauthoryear{Zhuo \bgroup \em et al.\egroup
  }{2011a}]{DBLP:conf/atal/ZhuoMY11}
Hankz~Hankui Zhuo, Hector Mu{\~{n}}oz{-}Avila, and Qiang Yang.
\newblock Learning action models for multi-agent planning.
\newblock In {\em Proceedings of the 10th International Conference on
  Autonomous Agents and Multiagent Systems {(AAMAS} 2011), Taipei, Taiwan, May
  2-6, 2011, Volume 1-3}, pages 217--224, 2011.

\bibitem[\protect\citeauthoryear{Zhuo \bgroup \em et al.\egroup
  }{2011b}]{DBLP:conf/aips/ZhuoYPL11}
Hankz~Hankui Zhuo, Qiang Yang, Rong Pan, and Lei Li.
\newblock Cross-domain action-model acquisition for planning via web search.
\newblock In {\em Proceedings of the 21st International Conference on Automated
  Planning and Scheduling, {ICAPS} 2011, Freiburg, Germany June 11-16, 2011},
  2011.

\bibitem[\protect\citeauthoryear{Zhuo \bgroup \em et al.\egroup
  }{2014}]{DBLP:journals/ai/ZhuoM014}
Hankz~Hankui Zhuo, H{\'{e}}ctor Mu{\~{n}}oz{-}Avila, and Qiang Yang.
\newblock Learning hierarchical task network domains from partially observed
  plan traces.
\newblock {\em Artif. Intell.}, 212:134--157, 2014.

\bibitem[\protect\citeauthoryear{Zhuo}{2015}]{DBLP:conf/aaai/Zhuo15}
Hankz~Hankui Zhuo.
\newblock Crowdsourced action-model acquisition for planning.
\newblock In {\em Proceedings of the Twenty-Ninth {AAAI} Conference on
  Artificial Intelligence, January 25-30, 2015, Austin, Texas, {USA.}}, pages
  3439--3446, 2015.

\end{thebibliography}

\end{document}